%% file: main.tex
\documentclass[10pt]{article} 

\usepackage[preprint]{tmlr}

\usepackage{hyperref}
\usepackage{url}
\usepackage[utf8]{inputenc} 
\usepackage[T1]{fontenc}    
\usepackage{xr-hyper}
\usepackage{booktabs}       
\usepackage{amsfonts}       
\usepackage{nicefrac}       
\usepackage{microtype}      
\usepackage{graphicx}
\usepackage{caption}
\usepackage{subcaption}
\usepackage{pgf}
\usepackage{lipsum}

\let\svthefootnote\thefootnote
\newcommand\freefootnote[1]{%
  \let\thefootnote\relax%
  \footnotetext{#1}%
  \let\thefootnote\svthefootnote%
}

\newcommand{\subfile}[1]{\input{#1}}
\usepackage{amsmath,amsthm,mathtools}
\usepackage{tikz}

\newtheorem{theorem}{Theorem}

\newtheorem{lemma}{Lemma}
\newtheorem{corollary}{Corollary}

\theoremstyle{definition}
\newtheorem{definition}{Definition}

\input{math_commands.tex}

\input{macros}

\title{Revisiting Parameter Sharing in Multi-Agent Deep Reinforcement Learning}

\author{\name J. K. Terry\textsuperscript{1,2,3} \email jkterry@umd.edu 
      \AND
      \name Nathaniel Grammel\textsuperscript{3} \email ngrammel@umd.edu
      \AND
      \name Sanghyun Son\textsuperscript{3} \email shh1295@umd.edu 
      \AND
      \name Benjamin Black\textsuperscript{2} \email benjamin.black@swarmlabs.com 
      \AND
      \name Aakriti Agrawal\textsuperscript{3} \email agrawal5@umd.edu
      }

\def\month{MM}  
\def\year{YYYY} 
\def\openreview{\url{https://openreview.net/forum?id=XXXX}} 

\begin{document}

\maketitle

\begin{abstract}
Parameter sharing\freefootnote{\textsuperscript{1} Corresponding author}\freefootnote{\textsuperscript{2} Swarm Labs}\freefootnote{\textsuperscript{3} Department of Computer Science, University of Maryland, College Park}, where each agent independently learns a policy with fully shared parameters between all policies, is a popular baseline method for multi-agent deep reinforcement learning. Unfortunately, since all agents share the same policy network, they cannot learn different policies or tasks. This issue has been circumvented experimentally by adding an agent-specific indicator signal to observations, which we term ``agent indication.'' Agent indication is limited, however, in that without modification it does not allow parameter sharing to be applied to environments where the action spaces and/or observation spaces are heterogeneous. This work formalizes the notion of agent indication and proves that it enables convergence to optimal policies for the first time. Next, we formally introduce methods to extend parameter sharing to learning in heterogeneous observation and action spaces, and prove that these methods allow for convergence to optimal policies. Finally, we experimentally confirm that the methods we introduce function empirically, and conduct a wide array of experiments studying the empirical efficacy of many different agent indication schemes for image based observation spaces.
\end{abstract}

\section{Introduction}

Reinforcement Learning (RL) is the intersection of machine learning and optimal control. It allows an agent in an environment to learn a policy which output actions given observations in order to maximize a reward signal defined by the environment. However, many real-life scenarios are more accurately modeled by multi-agent reinforcement learning (MARL), where multiple agents act and learn simultaneously.

MARL methods are often further subdivided according to two main properties: whether the agents are cooperative or competitive (or mixed), and if various components are centralized or decentralized. In decentralized systems, each agent takes decisions and learns independently, without access to other agents' observations, actions, or policies. This is similar to real-life scenarios like groups of animals, as opposed to a `hive-mind' controller acting as a single large RL agent (the centralized case). However, decentralized learning is intuitively more difficult, as information must be independently learned by each agent or passed between them. Therefore, most modern MARL research follows the paradigm of Centralized Training, Decentralized Execution (CTDE) \citep{lowe2017multi}, where agents execute their individual policies separately, but have access to other agents' observations during training. This paper specifically considers parameter sharing (which we refer to as ``parameter sharing''), a CTDE execution scheme where all policies are represented by a single neural network with the same shared parameters for each. 



Because parameter sharing has only one policy network or function, it is easy to assume that it can only handle agents with identical behavior in the environment, which is a significant limitation. This assumption has been relaxed for different types of agents, by adding an indication of the observing agent to the observations, allowing a single policy to serve multiple agents~\citep{foerster2016learning,gupta2017cooperative}. While this is known to work empirically, our work is first to formally define ``agent indication'' and prove that it allows convergence to optimal policies (\autoref{sec:3.2}). 

Even with agent indication method parameter sharing suffer from another limitation. With parameter sharing the agents cannot have different action and observation spaces with a neural network that can only accept and produce fixed sized tensors. This work's second contribution is formally introducing two padding based methods that are capable of resolving this for the first time, and similarly proving that these methods allow for convergence to optimal policies (\autoref{sec:3.2}, \autoref{sec:3.3}). We additionally provide experimental validation for these methods in~\autoref{sec:experiments}.

One implementation detail that comes up when using agent indication is the specific method you indicate an agent in the observation space with. There are two ``natural'' methods to do this with vector observation spaces that have been used in the literature, however no existing literature applies this to image based observation spaces. This work poses five simple methods for agent indication in image based observation spaces, and conducts experiments based on hyperparameter tuning sweeps to roughly determine the relative performance of these methods, and in doing so gain initial insight into the likely sensitivity environments have to these methods and if any of these methods may be more universally useful than the others.

These experiments additionally experimentally confirm that the methods we theoretically describe function experimentally, as one environment we ran tests on required applying every method we describe in order to be able to learn.




\section{Background and Related Work}





\subsection{Partially-Observable Stochastic Games}

In multi-agent environments,
MDPs can be extended to include a set of actions for each agent to create the Multi-agent MDP (``MMDP'') model~\citep{boutilier1996planning}. However, this model assumes all agents receive the same rewards. The Stochastic Games model (sometimes called \emph{Markov Games}), introduced by~\citep{shapley1953stochasticgames}, extends this by allowing a unique reward function for each agent. The Partially-Observable Stochastic Games (``POSG'') model, defined below, extends the Stochastic Games model to settings where the state is only partially observable (akin to a POMDP); it can be seen as equivalent to a kind of extensive game~\citep{Kuhn1953ExtensiveGames}. This is a more general model, and what we use in this paper. We define this model for our later use in \autoref{def:posg}. A good in depth overview of the POSG model and intuition can be found in \citet{terry2020pettingzoo}.

\begin{definition}[Partially-Observable Stochastic Game]
\label{def:posg}
  A \emph{Partially-Observable Sto\-chas\-tic Game} (POSG) is a tuple
  $\langle \states, \aN, \set{\actions_{i}}
  ,\transition, \set{\reward_{i}}
  ,
  \set{\observations_{i}}
  ,
  \set{\obsfunc_{i}}
  \rangle$, where:
  \begin{itemize}
  \item $\states$ is the set of possible \emph{states}.
  \item $\aN$ is the \emph{number of agents}. The \emph{set of
      agents} is $\agentset$.
  \item $\actions_{i}$ is the set of possible \emph{actions} for agent
    $i$. We denote by $U = {\prod_{i\in\agentset} 
    \actions_{i}}$ the set of possible \emph{joint actions} over all agents.
  \item
    $\transition\colon {{\states \times 
    {\prod_{i\in\agentset} 
    \actions_{i}}
    \times \states}  \to 
    [0,1] }$ is the (stochastic)
    \emph{transition function}. 
  \item
    $\reward_{i}\colon \states \times
        {\prod_{i\in\agentset} 
    \actions_{i}}
    \times \states \to
    \R$ is the \emph{reward function} for agent $i$.
  \item $\observations_{i}$ is the set of possible \emph{observations}
    for agent $i$.
  \item
    $\obsfunc_{i}\colon {{\states \times \actions_{i} \times
    \observations_{i}} \to [0,1]}$ is the \emph{observation
      function}. 
  \end{itemize}
\end{definition}


\subsection{Centralised Training Decentralised Execution}
Most modern MARL research follows the paradigm of Centralized Training, Decentralized Execution (CTDE) \citep{lowe2017multi}, where agents execute their individual policies separately, but have access to other agents' observations during training. One like of work is value decomposition based method (QMIX~\citep{rashid2018qmix}, FACMAC~\citep{facmac}) where a centralised value function, usually a joint Q-function, is decomposed into local utility function. Another line of work is based on multi-agent policy gradient (MADDPG~\citep{lowe2017multi}) where a centralised value function is usually learned to evaluate current joint policy and guide the update of each local policy.  
\subsection{Parameter Sharing}
Parameter sharing is important to reduce the computational cost of separate policy or action-value function for multiple agents and hence commonly used with CTDE methods. Full parameter sharing (which we refer to as ``parameter sharing''), where all policies are represented by a single neural network with the same shared parameters was first introduced by~\citet{tan1993multi} for classical RL and later concurrently introduced to cooperative multi-agent \emph{deep} reinforcement learning by~\citet{foerster2016learning, gupta2017cooperative,chu2017parameter}. This simple method has since been used to remarkable efficacy in various applications, such as~\citep{zheng2018magent, yu2022dinno, chen2021deep, yang2018mean}. 
It is worth noting that parameter sharing is equivalent to naive self-play in competitive environments.

\subsection{Coping with Heterogeneity}

Agent indication for parameter sharing was first implemented in~\citet{foerster2016learning}, and has been used in numerous derivative works. We are aware of no work studying the best method of this in image based observations, theoretically studying methods for this, or attempting to cope with heterogeneous action or observation spaces outside of padding the action space of medivac unit in \citet{samvelyan2019starcraft}.
In the case of the medivac, actions greater than 5 specify a target allied agent index to heal, as opposed to combat agents, where actions greater than 5 specify an enemy agent index to attack. This overloading of the same range of the action space to do different things to different sets of agents is effectively equivalent to action padding. However, the authors do not justify this choice of action encoding or comment on it in their paper. 

{There are other ways of coping with heterogeneity as well like independent learning (IPPO \citet{ippo}, IQL \citet{iql}) or completely decentralised learning. However, these methods do not scale well with the increasing number of agents. In a practical scenario methods like parameter sharing are computationally efficient.}

\section{Theoretical Results} \label{sec:heterogagents}
In this section, we call the agents of a POSG \emph{homogeneous} if
they can be reordered into any other order without changing the behavior of the POSG. Formally we chose to define this as meaning all agents have identical
action spaces, observation spaces, and reward functions,
and that the transition function is symmetric with respect
to permutations on the actions input. If a POSG is not homogeneous, 
we definite it to be heterogeneous.

Parameter sharing has been traditionally seen as a technique that can
only be used in games with homogeneous agents because of the fact that
a single neural network is learned and shared among all
agents. However, in this section we present a novel method of modifying
the observation spaces, and prove that it can allow for the use of parameter sharing in
the case of heterogeneous agents.

\subsection{Disjoint Observation Spaces Allows for Learning an Optimal Policy}

When agents are not homogeneous, it is not clear that parameter
sharing can be applied. However, we note that a single policy can be
used in cases where the observation spaces of the agents are
disjoint. We state this claim and prove it, as a preliminary for the full ``agent indication'' technique described in the following section.
\begin{lemma} \label{lem:hetdisj}
  If
  $G = \langle \states, \aN, \set{\actions_{i}}, P, \set{\reward_{i}},
  \set{\observations_{i}}, \set{\obsfunc_{i}} \rangle$ is a POSG such
  that $\set{\observations_{i}}_{i\in\agentset}$ is disjoint (i.e.,
  $\observations_{i}\cap \observations_{j} = \emptyset$ for all $i\ne j$), then any
  collection of policies $\set{\pol_{i}}_{i\in\agentset}$ can be
  expressed as a single policy
  $\pol^{\agentset}\colon
  \left(\bigcup_{i\in\agentset}\observations_{i}\right) \times
  \left(\bigcup_{i\in\agentset}\actions_{i}\right)\to[0,1]$ which,
  from the perspective of any single agent $i$, specifies a policy
  equivalent to $\pol_{i}$.\footnote{Formally, for any agent
    $i\in\agentset$, observation $\observation\in\observations_{i}$,
    and action $a\in\actions_{i}$,
    $\pol^{\agentset}(\observation, a) = \pol_{i}(\observation, a)$.}
\end{lemma}
\begin{proof}
  Let $\observations = \bigcup_{i\in\agentset} \observations_{i}$ be
  the set of all observations across all agents, and similarly define
  $\actions = \bigcup_{i\in\agentset}\actions_{i}$ to be the set of
  all actions available to agents. Note that while $\observations$ is
  a union of $\aN$ disjoint sets, it is not necessarily true that
  $\set{\actions_{i}}_{i\in\agentset}$ is disjoint and so $\actions$
  is not necessarily the union of disjoint sets.

  Define $\iota\colon \observations \to \agentset$ as
  follows: $\iota(\observation)$ is the (unique) agent
  $i$ for which $\observation\in\observations_{i}$. Thus, for all
  $\observation\in\observations$, we have that
  $\observation\in\observations_{\iota(\observation)}$. Note
  that $\iota$ is well-defined specifically because the
  observation sets are disjoint, and thus each observation
  $\observation\in\observations$ appears in exactly one agent's
  observation space.
  
  Now, we define our single policy
  $\pol^{\agentset} \colon\observations\times\actions\to[0,1]$.
  Let
  \vspace{-1.5ex}
  \begin{equation} \label{eq:singlepol}
    \pol^{\agentset}(\observation, a) = 
    \begin{cases}
      \pol_{\iota(\observation)}(\observation, a) & \text{if }a\in\actions_{\iota(\observation)} \\
      0 & \text{otherwise}
  \end{cases}
  \end{equation}
  One can see from this definition that for any agent $i\in\agentset$,
  for any $\observation\in\observations_{i}$, and for any
  $a\in\actions_{i}$, we have
  $\pol^{\agentset}(\observation, a) = \pol_{i}(\observation,
  a)$. Thus, from the view of agent $i$, $\pol^{\agentset}$ defines a
  policy consistent with its own policy $\pol_{i}$.
\end{proof}
\begin{corollary} \label{cor:hetopt}
  For any Partially-Observable Stochastic Game
  $G = \langle \states, \aN, \set{\actions_{i}}, P, \set{\reward_{i}},
  \set{\observations_{i}}, \set{\obsfunc_{i}} \rangle$ with disjoint
  observation spaces {for which optimal individual policies exist for each agent}, there exists a {joint} policy
  $\pol^{*}\colon { {(\bigcup_{i\in\agentset}\observations_{i})} \times
   {(\bigcup_{i\in\agentset}\actions_{i})}} \to [0,1]$ which is optimal for
  all agents; i.e.\ 
  $\forall {i\in\agentset}, \observation\in\observations_{i},
  a\in\actions_{i}$, we have 
  $\pol^{*}(\observation, a) = \pol_{i}^{*}(\observation, a)$, where
  $\pol_{i}^{*}$ is an optimal individual policy for agent $i$.
\end{corollary}

To briefly recap the intuition of the method we propose in this section---if by any means, the identity of an agent is indicated in the observation space passed to a full parameter sharing based learning method during training, then the policy will be able to learn to distinguish each agent and act in a manner adapted to that agent specifically, despite only being a single network.

\subsection{Agent Indication Allows for Representing Optimal Policies}
\label{sec:3.2}
When observation spaces are not disjoint---that is, the raw observations from the environment do not indicate the identity of the agent---we can force them to be disjoint by ``tagging'' the observations with an identifier unique to each agent. This technique is referred to as ``agent indication'' and is described formally below.
\begin{theorem} \label{thm:hetnondisj}
  For every POSG, there is an equivalent POSG with disjoint
  observation spaces.
\end{theorem}
\begin{proof}
  Let
  $G = \langle \states, \aN, \set{\actions_{i}}, P, \set{\reward_{i}},
  \set{\observations_{i}}, \set{\obsfunc_{i}} \rangle$ be a POSG with
  non-disjoint observation spaces. We define
  $G' = \langle \states, \aN, \set{\actions_{i}}, P,
  \set{\reward_{i}}, \set{\observations'_{i}}, \set{\obsfunc'_{i}}
  \rangle$, where $\observations'_{i}$ and $\obsfunc'_{i}$ are derived
  from $\observations_{i}$ and $\obsfunc_{i}$ respectively, as
  described below.

  For each agent $i$, we define
  $\observations'_{i} = \observations_{i} \times \set{i} =
  \set{(\observation, i) \mid
    \observation\in\observations_{i}}$. Intuitively, we ``attach''
  information about the agent $i$ to the observation. Now, for each
  agent $i\in\agentset$, we define
  $\obsfunc'_{i}\colon \actions_{i} \times \states \times
  \observations'_{i} \to [0,1]$ as
  $\obsfunc'_{i}(a,s,(\observation,i)) =
  \obsfunc_{i}(a,s,\observation)$. This is equivalent to $G$ in the
  sense that there is a family of bijections
  $f_{i}\colon\observations_{i}\to\observations'_{i}$ such that
  $\forall i\in\agentset, \forall a\in\actions_{i}, \forall
  s\in\states, \forall \observation\in\observations_{i}$,
  $\obsfunc_{i}(a,s,\observation) =
  \obsfunc'_{i}(a,s,f_{i}(\observation))$ (specifically,
  $f_{i}(\observation) = (\observation, i)$).
\end{proof}
\autoref{thm:hetnondisj} together with 
\autoref{cor:hetopt} shows that an optimal single 
policy $\pol^{*}\colon (\bigsqcup_{i\in\agentset} 
\observations_{i}) \times (\bigcup_{i\in\agentset} 
\actions_{i})\to [0,1]$\footnote{$\bigsqcup_{i}\observations_{i}$ 
is the \emph{disjoint union}: $\bigsqcup_{i
\in\agentset}\observations_{i} := \bigcup_{i\in\agentset} (\observations_{i} \times \set{i}) = \bigcup_{i\in\agentset}\set{(\observation, i) \mid \observation\in\observations_i}$.} 
is consistent with the optimal policies of each 
agent: if $\pol_{i}^{*}\colon
\observations_i\times\actions_{i}\to[0,1]$ is
an optimal individual policy for 
agent $i$, then $\pol^{*}((\observation, i), a) = \pol_{i}^{*}(\observation, a)$ for every action
$a\in\actions_{i}$.

In practice, many environments lend themselves to 
disjoint observation spaces (e.g., games with a third-person point of view), and \autoref{lem:hetdisj}
shows that one can use a single algorithm
(e.g., a single neural network) to learn a single
 policy that behaves differently for each agent. 
Additionally, \autoref{thm:hetnondisj} says that for 
non-disjoint observation spaces, we can
``attach'' the identity of each agent to its observations
(e.g.\ by superimposing an identifier that is distinct for each agent
onto the observations before they are input to the learning algorithm),
forcing the modified observations to be elements of disjoint
observation spaces so that \autoref{lem:hetdisj}
applies. \autoref{cor:hetopt} says that there is no
disadvantage to this approach in terms of the optimality of
the learned policy. In situations where the representations
of the observations of each agent do not have the 
same size, they can all be ``padded'' to the size 
of the largest and learning can proceed as normal. In other words, if you add 0s (or a similar value) to pad observation tensors with a smaller shape than others such that they are the same size, then this can be passed to a policy network and it can learn as normal. This padding allows neural networks or other policies of fixed dimension input to control agents with differing observation spaces.

\subsection{Padding Heterogeneous Action Spaces Allows for Representing Optimal Policies}
\label{sec:3.3}
Consider an environment where each agent 
$i\in\agentset$ has a different action space 
$\actions_{i}$ and, critically, that for some
pair of agents $i,i'\in\agentset$ we may have
$|\actions_{i}|\ne |\actions_{i'}|$. In the formal
model, this is not a problem, but it does present a
minor issue in implementation. Specifically, 
suppose we have a learning algorithm that has 
learned a policy $\pol\colon \observation 
\times \actions \to [0,1]$ for a single agent.
In practice, the learning algorithm,
given an observation $\observation$, outputs the 
induced probability distribution 
$\pol(\observation, \cdot)$ often as a vector 
$\vec{a} \in [0,1]^{|\actions|}$  with 
$\ell_{1}$-norm $1$.

If each agent's behavior is learned as a separate 
policy, the learning algorithm for agent $i$ will 
output a probability vector in 
$[0,1]^{|\actions_{i}|}$. However, if there are two
agents $i,i'$ with $|\actions_{i}| \ne |\actions_{i'}|$,
using the same network for both agents $i$ 
and $i'$ appears to preclude the use of parameter sharing, as the output vectors
have different dimensions.

We can address this as follows. Suppose we wish for
our algorithm to learn a single policy 
$\pol^{\agentset}$ for all agents, as in the 
previous section. One option is to have the algorithm 
output a probability vector in $[0,1]^{|\actions|}$, 
where $\actions = \bigcup_{i\in\agentset} \actions_{i}$
as before. Now, when the algorithm outputs a vector
$\vec{a}\in[0,1]^{\actions}$ for agent $i$, we can 
simply ``clip'' the vector and consider only the 
subvector corresponding to actions in $\actions_{i}$.
We can write this formally as $\vec{a}_{\actions_{i}}
\in [0,1]^{|\actions_{i}|}$. 

This can be quite space-inefficient, though, as in the 
worst case (when all agents have disjoint action 
spaces) $|\actions| = \sum_{i\in\agentset} \actions_{i}$.
Further, since an agent $i$ will never perform an
action $a\notin \actions_{i}$, much space is wasted 
representing the output as a sparse vector (i.e.\ one with 
many zeros). A more practical alternative is to
instead output a vector whose dimension is only as large
as the largest action space, padding the vector with
zeros for agents with smaller action spaces. Formally,
the learning algorithm can simply output a vector 
$\vec{a}\in[0,1]^{\alpha}$ for all agents, padding zeros
at the end where necessary. An agent $i$ with
$|\actions_{i}| < \alpha$ receiving a vector 
$\vec{a}\in[0,1]^{\alpha}$ can simply consider the 
subvector 
$\langle \vec{a}_{1}, \vec{a}_{2},\dots,\vec{a}_{|\actions_i|} \rangle$. 
Essentially, the learning algorithm pads the action 
vector to length $\alpha$, and each agent clips the vector
they receive to length $|\actions_{i}|$.

To briefly recap the intuition of the method we propose in this section---if you can ``pad'' action spaces so that a neural network is outputting actions of a fixed dimensionality and range, agents that can only accept less than this can either clip or throw out unneeded actions and the policy can still be represented. This padding allows neural networks or other policies of fixed dimension output to control agents with differing action spaces.

\section{Agent Indication Method}
{In this section we introduce the various agent indication methods for image based observations for heterogeneous agents.}

\subsection{Overview}

An important implementation detail while using agent indication is the exact method that is used to indicate the agent in the observations. In the case of 1D vector observations, previous work have shown to it via one hot or binary encoding, depending on the type of environment. However in the case of image based observation spaces, which are very common, there are an incredibly large number of possible ways to indicate an agent. However, no prior works experimentally investigate the impact of these methods on the end performance of common deep reinforcement learning regimes.

Thus, in this section we proposes five simple approaches to agent indication for image based observations (there are no past works to draw from), specifically trying to pick the simplest approaches we were able to think of. To evaluate the relative success of these approaches, we selected five interesting image based environments (\autoref{env-selection}) and conducted large hyperparameter searches to find the best parameter shared PPO policy over each where the agent indication method was included as one hyperparameter. Each hyperparameter search included all the commonly searchers hyperparameters for PPO, detailed in \autoref{appendix-a}. We then retrained the best set of 10 hyperparameters 10 times each for each environment (5 for prospector), and present the expected reward of the policy during training averaged across all of the training runs.

Additionally, learning in the \emph{prospector} environment required using both agent indication and the two methods previously introduced for coping with action space and observation space heterogeneity, showing that the methods are able to empirically learn.

We caution that determining a universally best agent indication method is likely not possible and the success of any method will depend on various details in the environment and training regime. However, these experiments can give insight regarding how much the different methods may matter in practice and if a ``generally best'' agent indication method of those we postulate exists or if it tends to be very environment specific. This experimental design was chosen in order to mitigate the likely profound confounding impact other hyperparameters may have on the performance of a specific agent indication method.

\subsection{Proposed Agent Indication Methods}

\begin{figure}[h]
     \centering
     \begin{subfigure}[b]{\textwidth}
         \centering
         \includegraphics[trim=0 0 0 0, clip, width=0.47\textwidth]{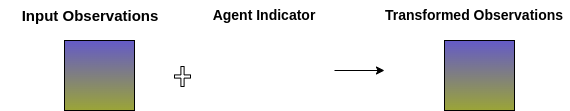}
         \rule{0.2px}{40px}
         \includegraphics[trim=0 0 0 0, clip, width=0.47\textwidth]{figures/identity.png}
    \caption{Identity Agent Indicator}
    \label{fig:in1}
    \end{subfigure}
    \vspace{3mm}

    \begin{subfigure}[b]{\textwidth}
         \centering
         \includegraphics[trim=0 0 0 0, clip, width=0.47\textwidth]{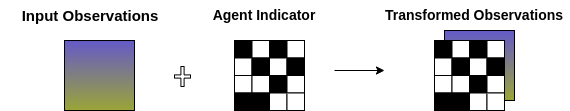}
         \rule{0.2px}{40px}
         \includegraphics[trim=0 0 0 0, clip, width=0.47\textwidth]{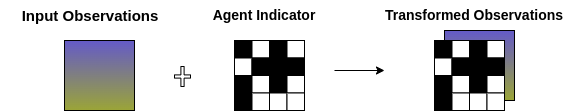}
    \caption{Geometric Agent Indicator}
    \label{fig:in2}
    \end{subfigure}
    \vspace{3mm}
     
    \begin{subfigure}[b]{\textwidth}
         \centering
         \includegraphics[trim=0 0 0 0, clip, width=0.47\textwidth]{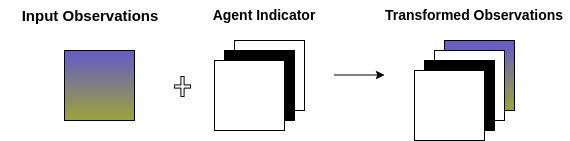}
         \rule{0.2px}{40px}
         \includegraphics[trim=0 0 0 0, clip, width=0.47\textwidth]{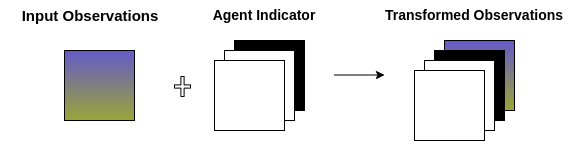}  
    \caption{Binary Agent Indicator}
    \label{fig:in3}
    \end{subfigure}
    \vspace{3mm}
    
    \begin{subfigure}[b]{\textwidth}
         \centering
         \includegraphics[trim=0 0 0 0, clip, width=0.47\textwidth]{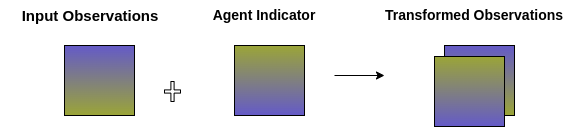}
         \rule{0.2px}{40px}
         \includegraphics[trim=0 0 0 0, clip, width=0.47\textwidth]{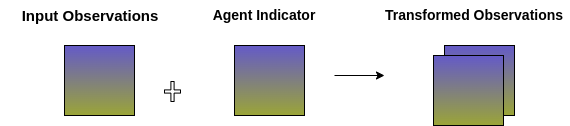}
    \caption{Inversion Agent Indicator}
    \label{fig:in4}
    \end{subfigure}
    \vspace{3mm}
     
     \begin{subfigure}[b]{\textwidth}
         \centering
         \includegraphics[trim=0 0 0 0, clip, width=0.47\textwidth]{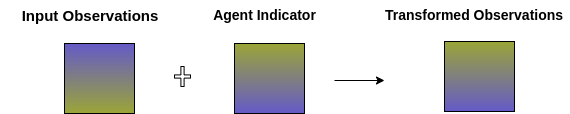}
         \rule{0.2px}{40px}
         \includegraphics[trim=0 0 0 0, clip, width=0.47\textwidth]{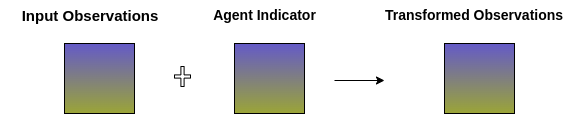}
    \caption{Inversion with replacement Agent Indicator}
    \label{fig:in5}
    \end{subfigure}

    \caption{The 5 different agent indicator methods for image based observations. We show two different agents on the left and right to understand the difference in the agent indicators for different methods.}
    \label{fig:agent_indicator}
\end{figure}

Per the reasoning described in the previous section, this work seeks to gain insight into the relative performance of the following five agent indication methods. {The input to the policy $\pi$ is the specific agent's observation $\omega \in \Omega_{i}$. We change these observations for the specific agent using the agent indication methods discussed below.} \autoref{fig:agent_indicator} shows the methods visually. 

\begin{enumerate}
  \item \emph{Identity} --- Do nothing to the observation, included for control. {The input to the policy $\pi$ is the specific agent's observation $\omega \in \Omega_{i}$. Thus, in this method two heterogeneous agents with different features and functions can have same observations but the policy will learn to produce the same output for them. Thus this method acts as a baseline.}
  \item \emph{Geometric} --- Add an additional channel with a per-pixel checkerboard for one type of agent and checkerboard translated by one pixel for the other type. {Thus, the input to the policy $\pi$ now has an additional channel added to the actual observation of the agent $\omega \in \Omega_{i}$. This additional channel is in the form of checkerboard as shown in figure \autoref{fig:agent_indicator} b, specific to the agent type.}
  \item \emph{Binary} --- Add additional channels, each of which is entirely black or white based on the type of an agent. {Thus, the input to the policy $\pi$ now has multiple additional channels added to the actual observation of the agent $\omega \in \Omega_{i}$. These additional channels are either completely white or black specific to the particular agent.}
  \item \emph{Inversion} --- Invert the color of the observation of certain types of agents and add it as a channel to the original observation. For agents that do not need inversion, duplicate original observation. {Thus, the input to the policy $\pi$ now has an additional channel added to the actual observation of the agent $\omega \in \Omega_{i}$. This additional channel is the inverted observation or actual observation as shown in \autoref{fig:agent_indicator} d. Note here that Inversion method will work only with environments with two kinds of agents.}
  \item \emph{Inversion with Replacement} --- The same as \emph{Inversion}, but the inverted observation (or the same duplicate observation for agents that do not need it) is used in place of the original observation. {Similar to Inversion this method will only work with environments with two kinds of agents.}
\end{enumerate}

\section{Experiments}
\label{sec:experiments}

{In this section we are trying to test the above methods on the 5 different MARL environment. We show that the policy converges with these agent indication methods and also give insights on which method works well with which environment.}

\subsection{Environment Selection}
\label{env-selection}

We selected five different environments for testing our work. They were selected from PettingZoo \citep{terry2020pettingzoo} library which has over 60 diverse multi-agent reinforcement learning environments with a standardized API. PettingZoo has two ``classes'' of environments with image based observations- the ``Butterfly`` class with custom made games with pygame rendering and pymunk physics with a diverse set of rules, as well as the multi-player Atari games introduced in~\citet{terry2020multiplayer}. To maximize environment diversity, we chose every environment in the Butterfly class with heterogeneous agents (three), and two Atari games (the choice of which we will explain below).

The environments we choose are:
\begin{itemize}
  \item \emph{Cooperative Pong} --- A Butterfly environment inspired by the Atari pong environment, where two differently shaped pistons work together to keep the ball in the air for as long as possible.
  \item \emph{Knights Archers Zombies} --- A Butterfly environment where knight and archer agents collectively work together to prevent zombies from crossing the bottom of the screen.
  \item \emph{Prospector} --- A Butterfly environment where banker and prospector agents work together to collect gold, give to one another and deposit it in banks
  \item \emph{Pong} --- classic two player Atari pong, chosen due to it's well known status and as a simple test  competitive game
  \item \emph{Entombed Cooperative} --- an Atari game where two identical agents work together to progress to the bottom of the screen, chosen due to it's status as the only fully cooperative Atari game included in PettingZoo
\end{itemize}

More comprehensive documentation of all these environments is available at \url{pettingzoo.ml}.

All of the butterfly class environments are cooperative environments (which is when parameter sharing is conventionally described as being used), with heterogeneous types of agents. Entombed cooperative was chosen because it is the only cooperative multiplayer Atari game supported in PettingZoo (few were ever made), and Pong was chosen as a simple test of the impact of agent indication in a simple and very well known competitive environment. Note that the Atari environments have identical baseline observations for each agent (the pixels on the screen), making agent indication crucial to allow the policy to differentiate at all between the agents. All environments are pictured in \autoref{fig:env_photos}.

\subsection{Additional Implementation Details}



Prior to agent indication, each environment had the following preprocessing done to its observations using SuperSuit~\citep{SuperSuit2020}.

\begin{enumerate}
\item RGB images converted to grayscale using  \texttt{color\_reduction\_v0}
\item Image resized to 96x96 pixels via bi-linear interpolation with \texttt{resize\_v0}.
\item Stacked the last 4 frames along the channel dimension with \texttt{frame\_stack\_v1}.
\item In Knights Archers Zombies only, since individual agents can terminate early in this environment, and our training setup did not support this explicitly, we used \texttt{black\_death\_v2} to create an all black image to represent the terminal state for the dead agent.
\end{enumerate}

The following number of training timesteps for each environment were used:

\begin{itemize}
  \item \emph{Cooperative Pong} --- 4 million timesteps.
  \item \emph{Knights Archers Zombies} --- 10 million timesteps.
  \item \emph{Prospector} --- 100 million timesteps.
  \item \emph{Pong} --- 10 million timesteps.
  \item \emph{Entombed Cooperative} --- 10 million timesteps.
\end{itemize}

Additionally, the hyperparameter ranges we searched are described in Appendix~\ref{appendix-a}, and for the evaluation in the Pong environment, we evaluated our agent's performance by letting it play against the built-in bot implemented in the game's single player mode (as the rewards when playing against itself were not meaningful). All of our learning code was implemented by Stable Baselines 3 \citep{stable-baselines3}, and our hyperparameter search code was based upon RL Baselines3 Zoo \citep{rl-zoo3}. All code used in our experiments is available at \url{https://anonymous.4open.science/r/parameter-sharing-paper-EB2D/}.

\subsection{Results}

\begin{figure}[htp]
     \centering
     \begin{subfigure}[b]{0.3\textwidth}
         \centering
         \includegraphics[trim=0 0 0 0, clip, width=1.1\textwidth]{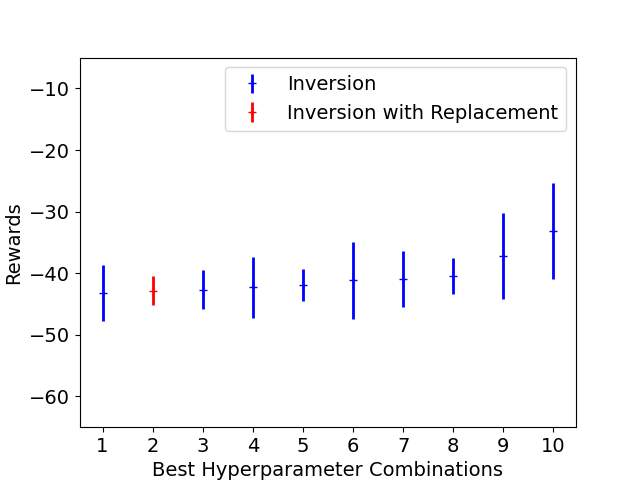}
         \caption{Cooperative Pong}
        \label{fig:1}
    \end{subfigure}
     \begin{subfigure}[b]{0.3\textwidth}
         \centering
         \includegraphics[trim=0 0 0 0, clip, width=1.1\textwidth]{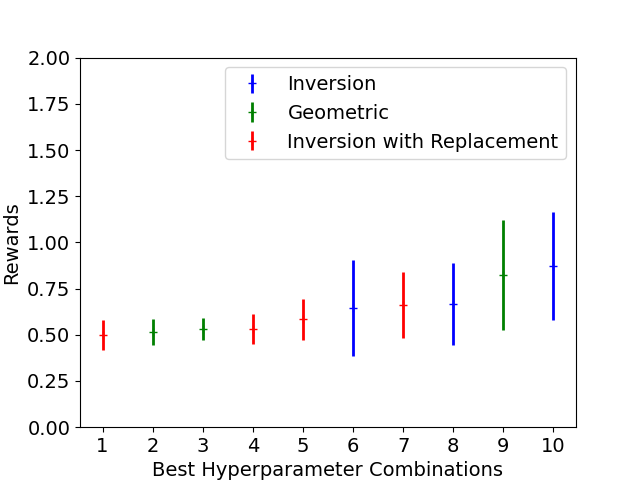}
         \caption{Knights Archers Zombies}
      \label{fig:2}
    \end{subfigure}
     \begin{subfigure}[b]{0.3\textwidth}
         \centering
         \includegraphics[trim=0 0 0 0, clip, width=1.1\textwidth]{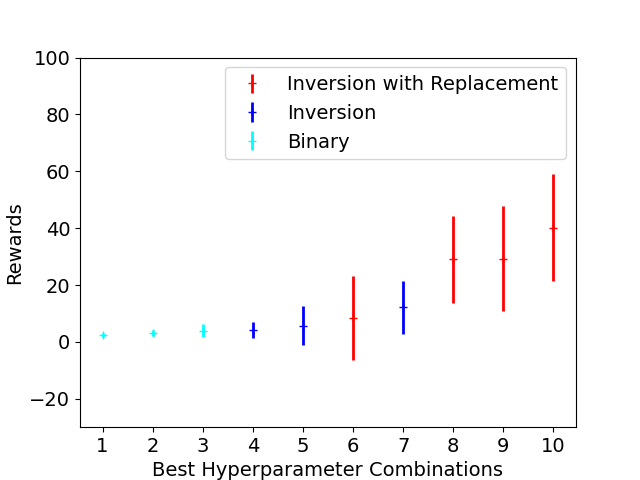}
         \caption{Prospector}
      \label{fig:3}
    \end{subfigure}
     \begin{subfigure}[b]{0.3\textwidth}
         \centering
         \includegraphics[trim=0 0 0 0, clip, width=1.1\textwidth]{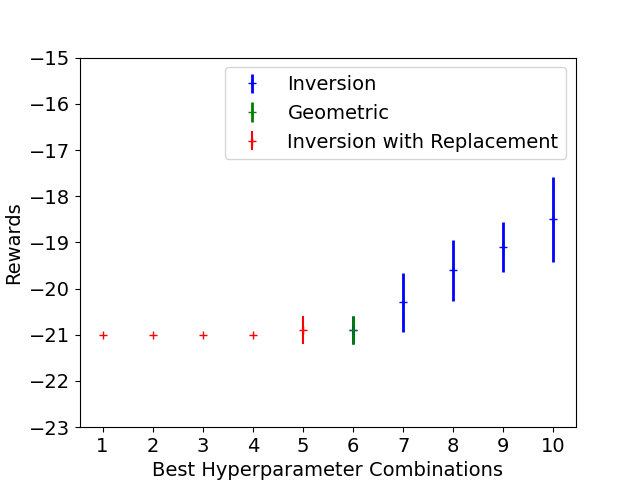}
         \caption{Pong}
      \label{fig:4}
    \end{subfigure}
     \begin{subfigure}[b]{0.3\textwidth}
         \centering
         \includegraphics[trim=0 0 0 0, clip, width=1.1\textwidth]{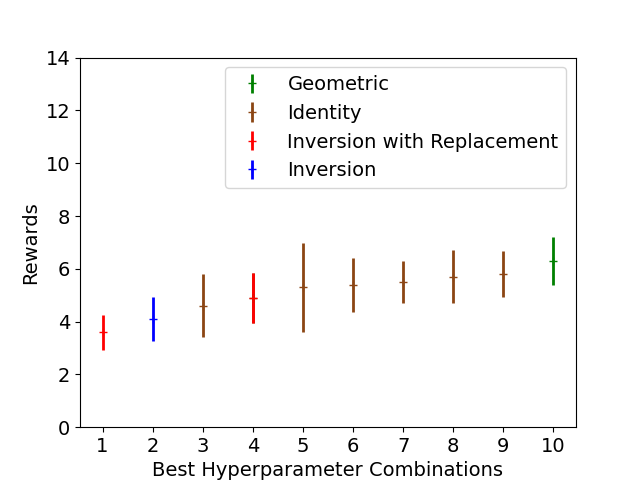}
         \caption{Entombed Cooperative}
      \label{fig:5}
    \end{subfigure}
     \begin{subfigure}[b]{0.3\textwidth}
         \centering
         \includegraphics[trim=0 0 85 10, clip, width=1.1\textwidth]{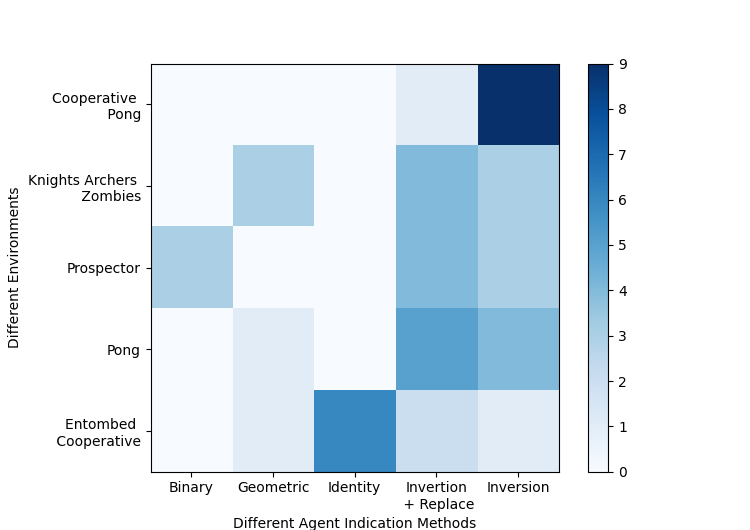}
         \caption{Heat-map}
      \label{fig:heatmap}
    \end{subfigure}

    \caption{(a)-(e) Error plots for different environment by picking the 10 best hyperparameters and agent indicator combination. (f) Heatmap showing the number of times a agent indication method happens to be one of the best method for a particular environment.}
    \label{fig:error_plot}
\end{figure}

The 10 best hyperpameters for each environment found during each automated hyperparameter search were all retrained 10 times (except for 5 in the case of prospector due to the longer run time). For each training process, the average-per-agent reward of the policy checkpoint were taken and averaged, in order to ensure that the rewards associated with hyperparameters are accurate and representative. These averaged-per-agent rewards are mentioned in \autoref{agent-ind-tables} and the average rewards for top 10 best hyperparameter/agent indication combination methods are mentioned in \autoref{agent-ind-tables} \autoref{fig:avg_reward}. We show the error plots for the 5 different environments with the 5 agent indication in \autoref{fig:error_plot}. Along with that we show heatmap of the number of times a agent indication method occurs in the top 10 hyperparameter search for each of the environments in \autoref{fig:heatmap}. This shows the likelihood of each method being the best.

In the results, the environments were found to be highly sensitive to agent indication method given the variance and specificity of which methods worked well in different environments. The most consistent methods across environments were \emph{Inversion} and \emph{Inversion with Replacement}, which were present in top 10 of all environments. Of the two, \emph{Inversion} performed better, being \emph{the} best for three out of five environments. The consistent performance is likely due to a combination of good differentiation between agents and good generalization possibilities for the simple color inverted features. As both the most consistent and highest performing method, \emph{Inversion} is likely a good initial choice for practitioners. 

The least consistent indicators were \emph{Identity} and \emph{Binary}, only appearing in the top 10 for a single environment each. \emph{Binary} did not perform well in any environment, only making an appearance in the top 10 in \emph{Prospector}. This is a somewhat surprising finding, as in theory, it provides sufficient information to differentiate agents. \emph{Identity} performs well only in the \emph{Entombed Cooperative} environment, perhaps because it has homogeneous agents unlike other environments and do not need parameter sharing.   

Another notable result is that \emph{Prospector}, an environment with heterogeneous action, observation and agents, was able to learn with parameter sharing using agent indication and the two padding methods this work introduces. This shows that the methods we introduce can empirically allow for learning.

Some more observations from the heat-map \autoref{fig:heatmap} are as follows. For \emph{Entombed Cooperative} environment, \emph{Identity} gave best results majority of the time. This is understandable because the agents are similar, in same setting and do not need different policies to achieve good results. However, geometric and inversion also were present in top 10, suggesting that additional information did not greatly impact the policy's ability to learn. For Cooperative Pong, \emph{Inversion} is clearly the best method. \emph{Binary} has shown good results only for the \emph{Prospector} environment where it still didn't give the best performance.

\section{Conclusion}
This paper introduces novel methods for coping with heterogeneous action and observation spaces in multi-agent environments learned via full parameter sharing, and shows that these and the previously known method of agent indication (for coping with heterogeneous agent behaviors) are able to represent optimal policies. We show that these methods are also capable of working together to empirically allow for learning.

We further offer experiments on the efficacy of different methods of agent indication in image based observation spaces, a previously unstudied problem, showing that it appears to be highly environment dependent, but that inverting the observations associated with one agent is likely a good baseline and starting point for researchers approaching new environments.



We hope these results will enable greater practical application of parameter sharing for multi-agent reinforcement learning. 


\bibliography{main}
\bibliographystyle{tmlr}

\newpage
\appendix

\label{Appendix}
\onecolumn

\section{Pseudo Code}

\section{Reward and Agent Indication Method Tables}
\label{agent-ind-tables}
\input{reward_tables}

\clearpage

\section{Hyperparameter Search Range}
\label{appendix-a}

\subsection{PPO}
\begin{table}[htp]
\caption{Hyperparameter search range for PPO algorithm. Continuous range is noted in square brackets.}
\label{hyperparameter-range-ppo}
\vskip 0.15in
\begin{center}
\begin{small}
\begin{sc}
\begin{tabular}{lcc}
\toprule
Hyperparameter & Range \\
\midrule
Batch size & 8, 16, 32, 64, 128, 256, 512 \\
Number of timesteps per update & 8, 16, 32, 64, 128, 256, 512, 1024, 2048 \\
Discount factor (Gamma) & 0.9, 0.95, 0.98, 0.99, 0.995, 0.999, 0.9999 \\
Entropy coefficient & [0.00000001, 0.1] \\
Clip range & 0.1, 0.2, 0.3, 0.4 \\
Number of epochs & 1, 5, 10, 20 \\
GAE coefficient (Lambda) & 0.8, 0.9, 0.92, 0.95, 0.98, 0.99, 1.0 \\
Maximum gradient norm & 0.3, 0.5, 0.6, 0.7, 0.8, 0.9, 1, 2, 5 \\
Value function coefficient & [0, 1] \\
Network architecture & (64, 64), (256, 256) \\
Activation function & tanh, relu \\
\bottomrule
\end{tabular}
\end{sc}
\end{small}
\end{center}
\vskip -0.1in
\end{table}

\clearpage

\section{Environment Images}

\begin{figure}[htp]
     \centering
     \begin{subfigure}[b]{0.3\textwidth}
         \centering
         \includegraphics[width=\textwidth]{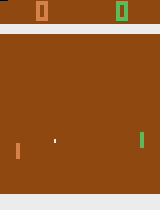}
         \caption{Pong}
        \label{fig:pong}
         \includegraphics[width=\textwidth]{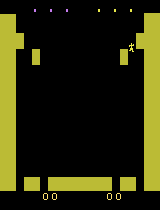}
         \caption{Entombed Cooperative}
        \label{fig:entombed_cooperative}
     \end{subfigure}
     \hfill
     \begin{subfigure}[b]{0.5\textwidth}
         \centering
         \includegraphics[width=0.9\textwidth]{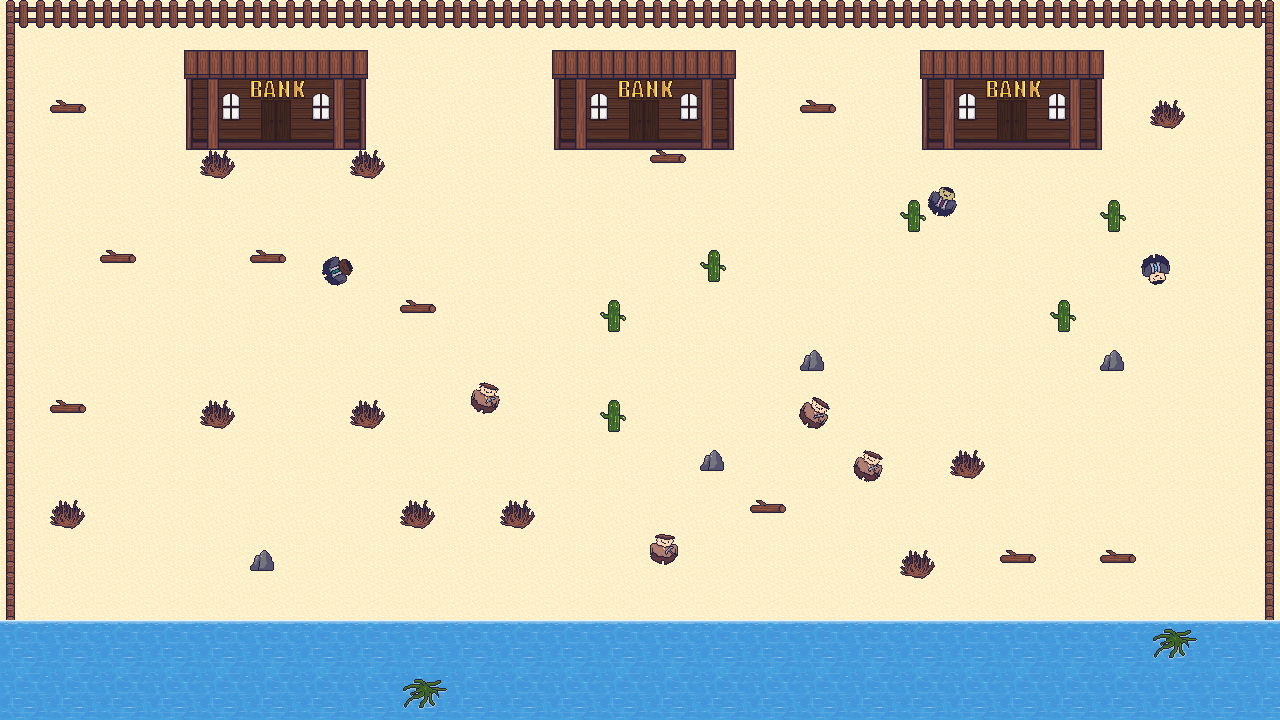}
         \caption{Prospector}
      \label{fig:prospector}
         \includegraphics[width=0.9\textwidth]{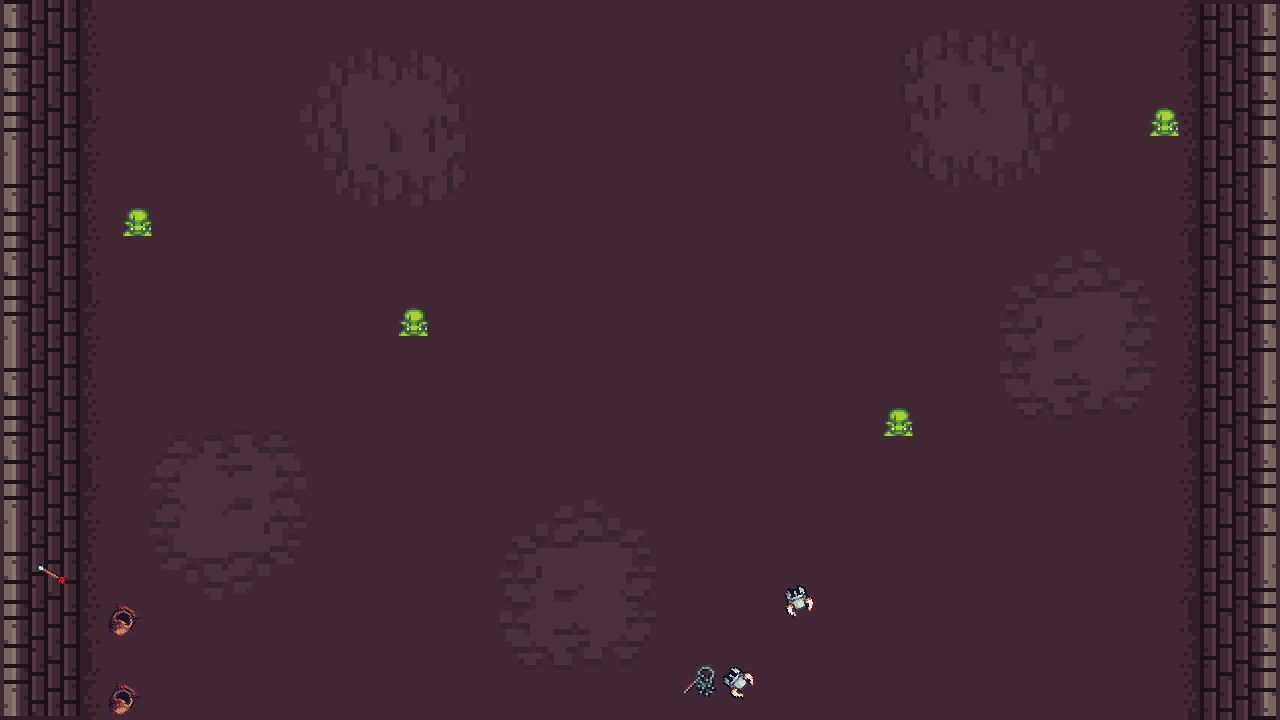}
         \caption{Knights Archers Zombies}
        \label{fig:kaz}
         \includegraphics[width=0.9\textwidth]{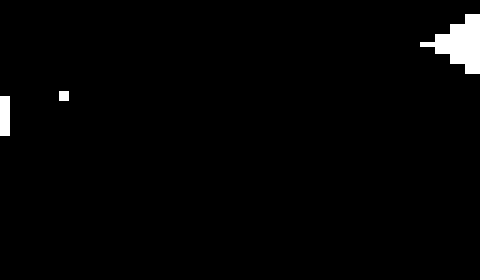}
         \caption{Cooperative Pong}
        \label{fig:cooperative_pong}
     \end{subfigure}
     \hfill

    \caption{Images of the benchmark environments from~\citet{terry2020pettingzoo}.}
    \label{fig:env_photos}
\end{figure}

\section{Learning graph for all environments for multiple hyper-parameters}

 \begin{figure}[!ht]
\centering
    \begin{subfigure}{0.45\linewidth}
        \includegraphics[width=\linewidth]{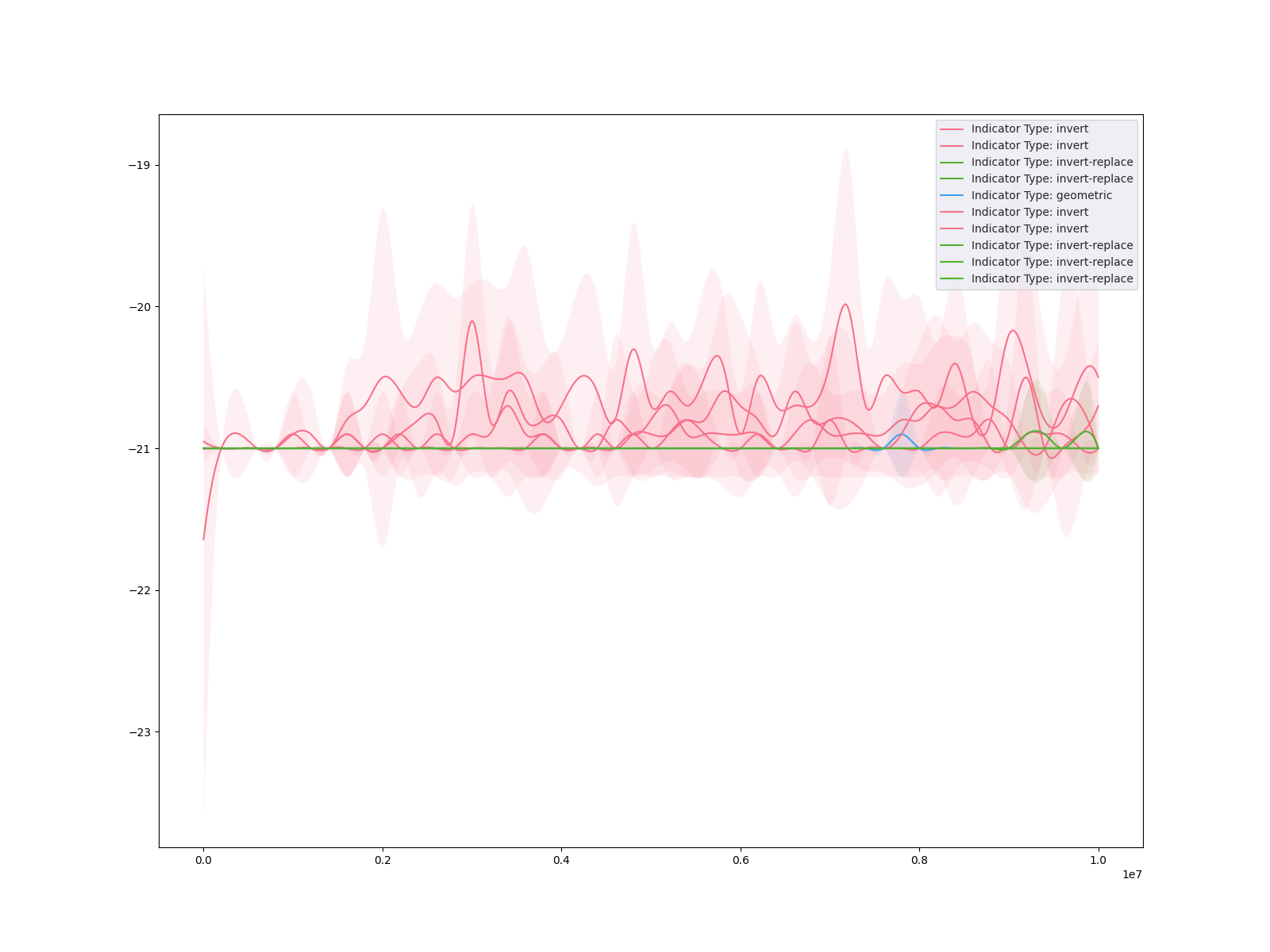}
    \caption{Pong}
    \end{subfigure}
    \hfil
    \begin{subfigure}{0.45\linewidth}
        \includegraphics[width=\linewidth]{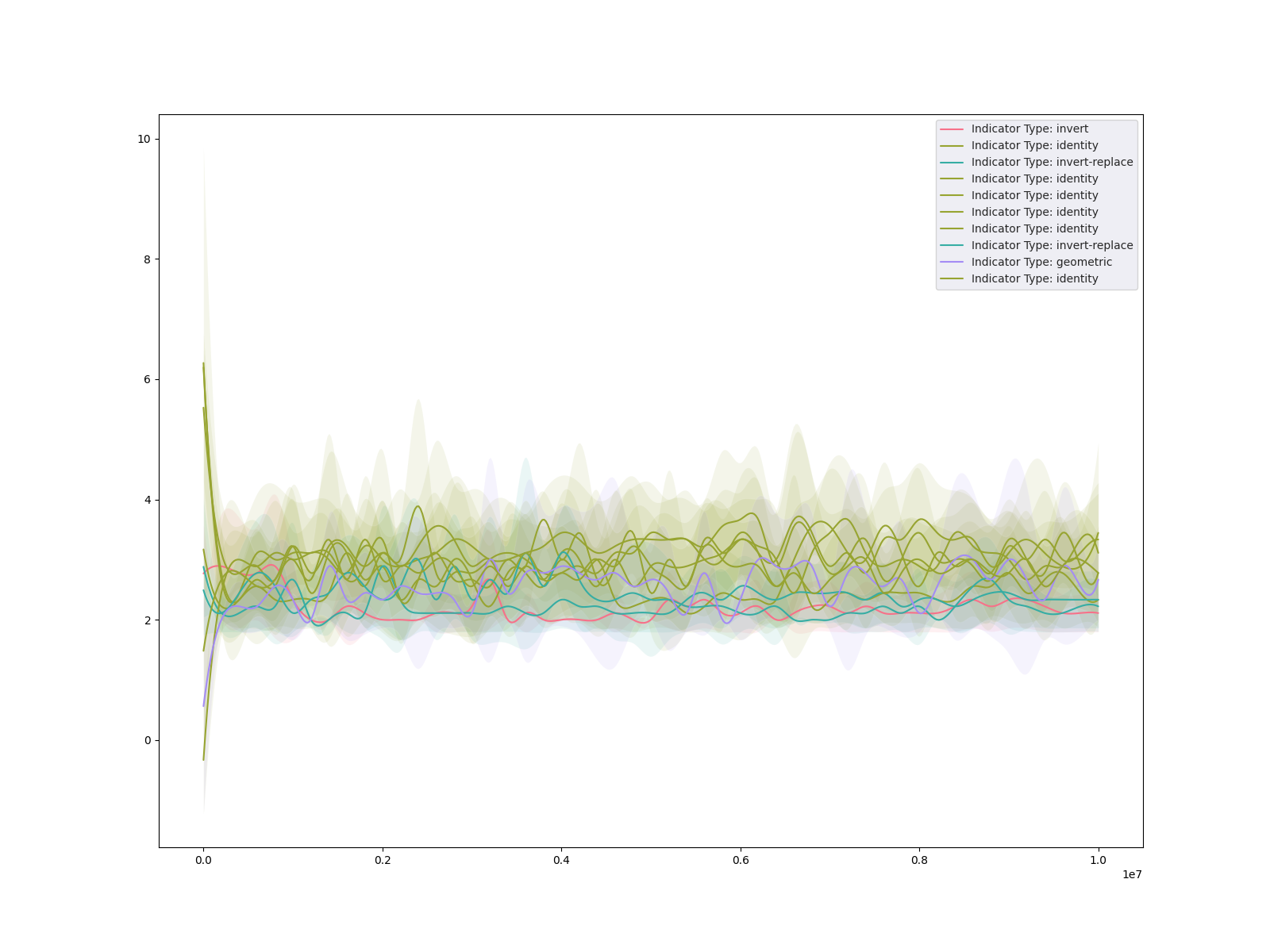}
    \caption{Entombed Cooperative}
    \end{subfigure}

    \begin{subfigure}{0.45\linewidth}
        \includegraphics[width=\linewidth]{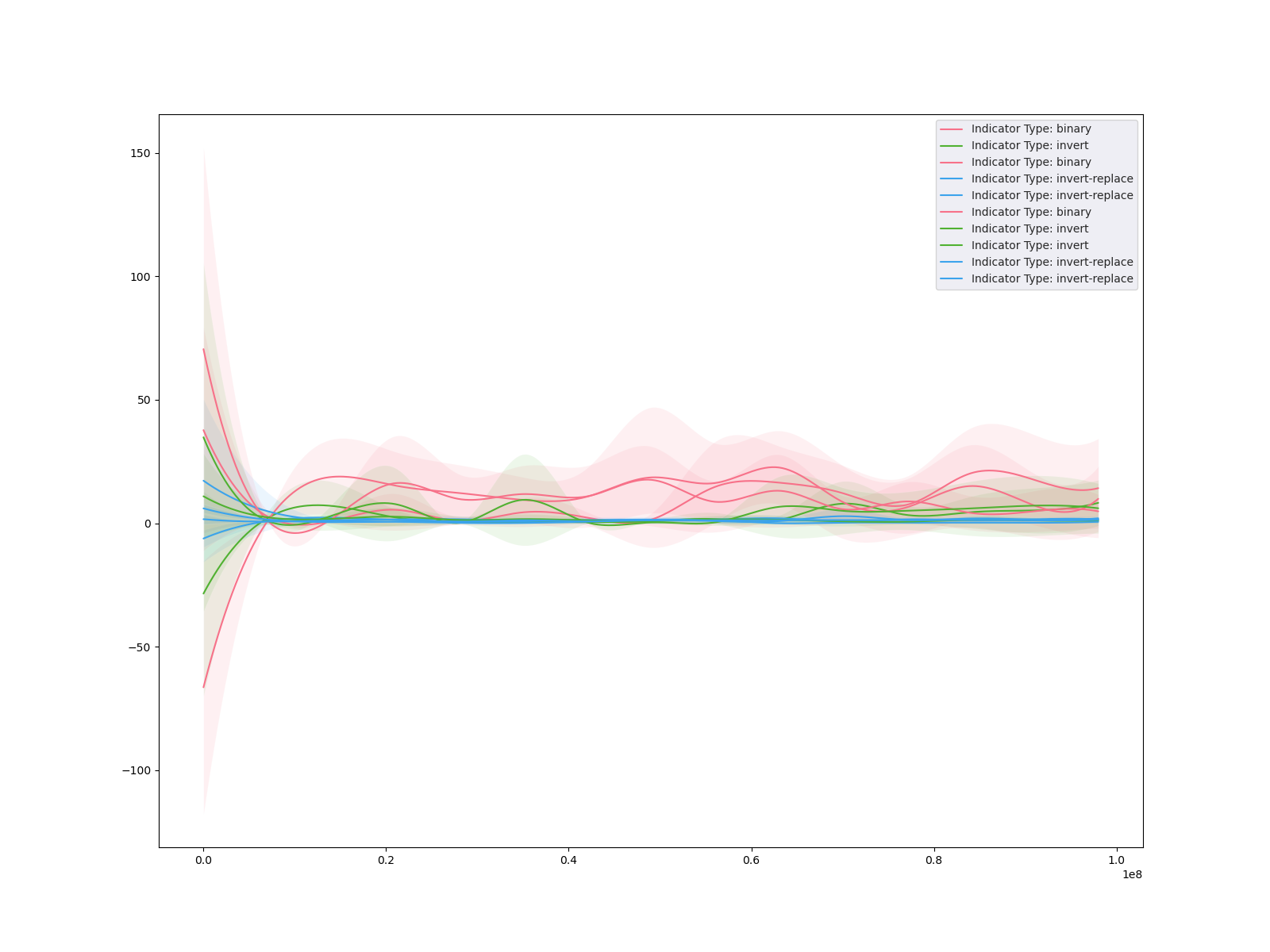}
    \caption{Prospector}
    \end{subfigure}
    \hfil
    \begin{subfigure}{0.45\linewidth}
        \includegraphics[width=\linewidth]{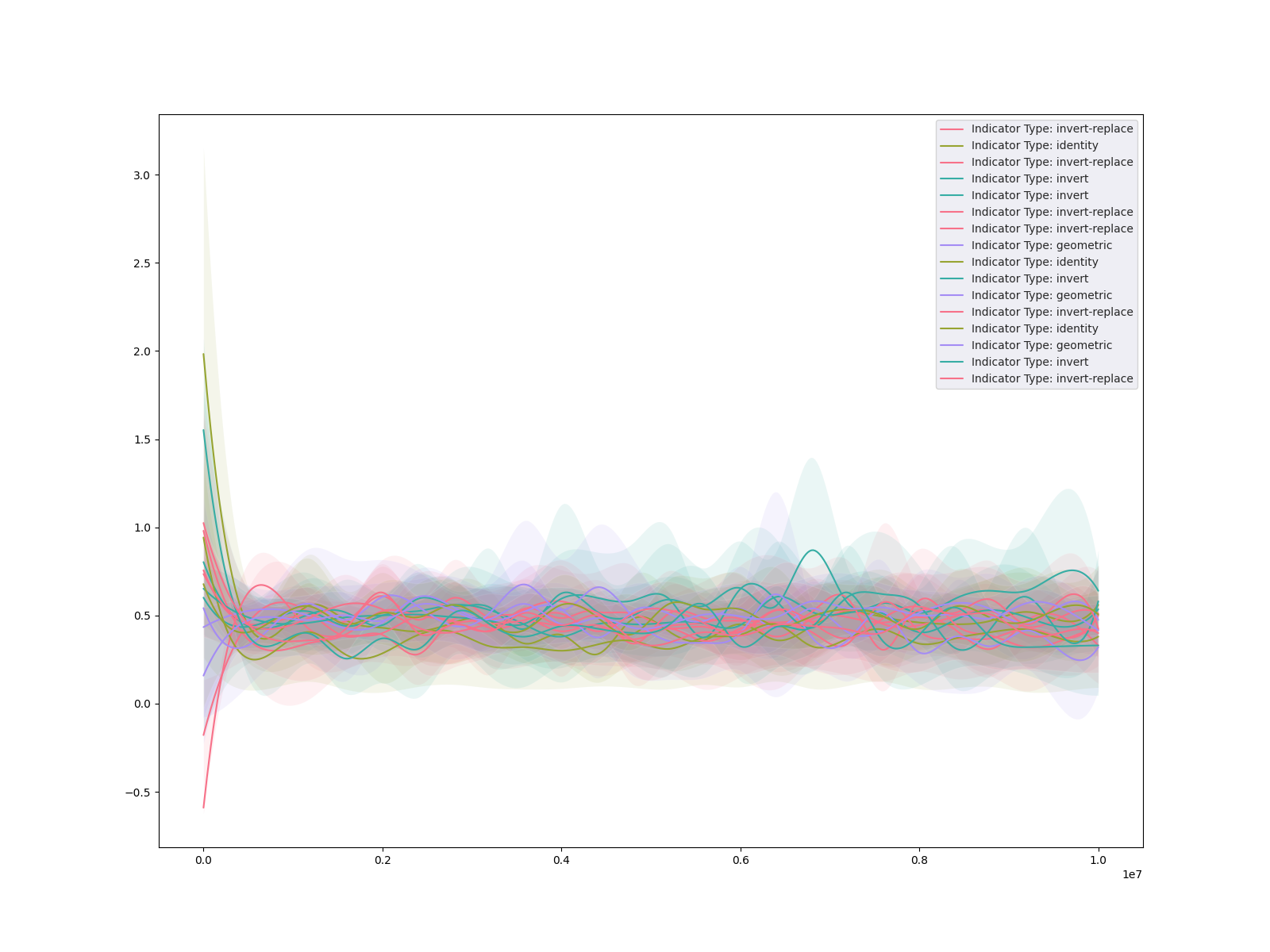}
    \caption{Knights Archers Zombies}
    \end{subfigure}

    \begin{subfigure}{0.45\linewidth}
        \includegraphics[width=\linewidth]{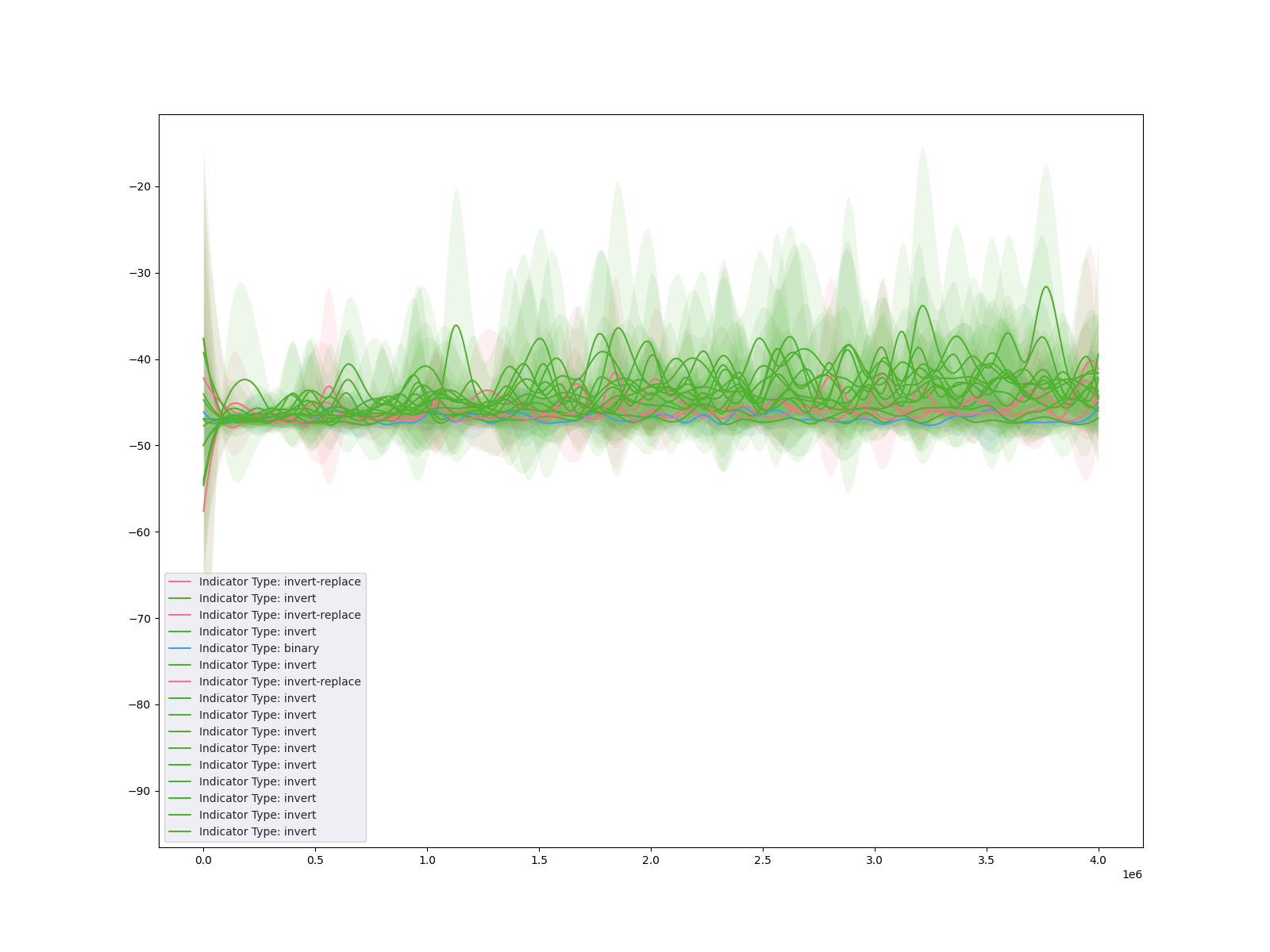}
    \caption{Cooperative Pong}
    \end{subfigure}
    
    \caption{Learning Graph}
    \label{fig:learning_graph}
    \end{figure}

\end{document}

%% file: math_commands.tex
\usepackage{amsmath,amsfonts,bm}

\def\eqref#1{equation~\ref{#1}}

\def\1{\bm{1}}

\DeclareMathAlphabet{\mathsfit}{\encodingdefault}{\sfdefault}{m}{sl}
\SetMathAlphabet{\mathsfit}{bold}{\encodingdefault}{\sfdefault}{bx}{n}

\newcommand{\R}{\mathbb{R}}



%% file: macros.tex
\newcommand{\states}{\mathcal{S}}
\newcommand{\aN}{N}
\newcommand{\agentset}{[\aN]}
\newcommand{\actions}{\mathcal{A}}

\newcommand{\transition}{P}
\newcommand{\reward}{R}
\newcommand{\observations}{\Omega}
\newcommand{\obsfunc}{O}
\newcommand{\observation}{\omega}
\DeclarePairedDelimiter{\set}{\{}{\}}
\newcommand{\pol}{\pi}

%% file: reward_tables.tex
\begin{table}[htp]
\caption{10 best agent indication methods for Cooperative Pong. 10 additional evaluations were done for each of the hyperparameters.}
\label{best-agent-indicator-cooperative-pong}
\vskip 0.15in
\begin{center}
\begin{small}
\begin{sc}
\begin{tabular}{cc}
\toprule
Agent indication method & Avg. Reward $\pm$ Std \\
\midrule
Inversion & -33.13 $\pm$ 7.8 \\
Inversion & -37.20 $\pm$ 6.92 \\
Inversion & -40.41 $\pm$ 2.91 \\
Inversion & -40.98 $\pm$ 4.52 \\
Inversion & -41.18 $\pm$ 6.31 \\
Inversion & -41.91 $\pm$ 2.55\\
Inversion & -42.31 $\pm$ 4.91 \\
Inversion & -42.66 $\pm$ 3.21\\
Inversion with Replacement & -42.85 $\pm$ 2.37\\
Inversion & -43.22 $\pm$ 4.5 \\
\bottomrule
\end{tabular}
\end{sc}
\end{small}
\end{center}
\vskip -0.1in
\end{table}

\begin{table}[htp]
\caption{10 best agent indication methods for KAZ. 10 additional evaluations were done for each of the hyperparameters.}
\label{best-agent-indicator-kaz}
\vskip 0.15in
\begin{center}
\begin{small}
\begin{sc}
\begin{tabular}{cc}
\toprule
Agent indication method & Avg. Reward \\
\midrule
Inversion & 0.87 $\pm$ 0.29 \\
Geometric & 0.82 $\pm$ 0.3\\
Inversion & 0.67 $\pm$ 0.22 \\
Inversion with Replacement & 0.66 $\pm$ 0.18\\
Inversion & 0.65 $\pm$ 0.26\\
Inversion with Replacement & 0.58 $\pm$ 0.11\\
Inversion with Replacement & 0.53 $\pm$ 0.06\\
Geometric & 0.53 $\pm$ 0.08\\
Geometric & 0.51 $\pm$ 0.07\\
Inversion with Replacement & 0.5 $\pm$ 0.08 \\
\bottomrule
\end{tabular}
\end{sc}
\end{small}
\end{center}
\vskip -0.1in
\end{table}

\begin{table}[htp]
\caption{10 best agent indication methods for Prospector. 5 additional evaluations were done for each of the hyperparameters.}
\label{best-agent-indicator-prospector}
\vskip 0.15in
\begin{center}
\begin{small}
\begin{sc}
\begin{tabular}{cc}
\toprule
Agent indication method & Avg. Reward \\
\midrule
Binary & 40.13 $\pm$ 18.76 \\
Binary & 29.33 $\pm$ 18.51\\
Binary & 29.09 $\pm$ 15.23\\
Inversion & 12.23 $\pm$ 9.36\\
Inversion & 8.46 $\pm$ 14.71\\
Inversion with Replacement & 5.75 $\pm$ 6.81\\
Inversion & 4.21 $\pm$ 2.89\\
Inversion with Replacement & 3.95 $\pm$ 2.18\\
Inversion with Replacement & 3.14 $\pm$ 1.02\\
Inversion with Replacement & 2.40 $\pm$ 0.73\\
\bottomrule
\end{tabular}
\end{sc}
\end{small}
\end{center}
\vskip -0.1in
\end{table}

\begin{table}[htp]
\caption{10 best agent indication methods for Pong. 10 additional evaluations were done for each of the hyperparameters.}
\label{best-agent-indicator-pong}
\vskip 0.15in
\begin{center}
\begin{small}
\begin{sc}
\begin{tabular}{cc}
\toprule
Agent indication method & Avg. Reward $\pm$ Std\\
\midrule
Inversion & -18.5 $\pm$ 0.92 \\
Inversion & -19.1 $\pm$ 0.54\\
Inversion & -19.6 $\pm$ 0.66\\
Inversion & -20.3 $\pm$ 0.64 \\
Geometric & -20.9 $\pm$ 0.3 \\
Inversion with Replacement & -20.9 $\pm$ 0.3 \\
Inversion with Replacement & -21.0 $\pm$ 0 \\
Inversion with Replacement & -21.0 $\pm$ 0 \\
Inversion with Replacement & -21.0 $\pm$ 0 \\
Inversion with Replacement & -21.0 $\pm$ 0\\
\bottomrule
\end{tabular}
\end{sc}
\end{small}
\end{center}
\vskip -0.1in
\end{table}

\begin{table}[htp]
\caption{10 best agent indication methods for Entombed Cooperative. 10 additional evaluations were done for each of the hyperparameters.}
\label{best-agent-indicator-entombed-cooperative}
\vskip 0.15in
\begin{center}
\begin{small}
\begin{sc}
\begin{tabular}{cc}
\toprule
Agent indication method & Avg. Reward \\
\midrule
Geometric & 6.3 $\pm$ 0.9 \\
Identity & 5.8 $\pm$ 0.87\\
Identity & 5.7 $\pm$ 1.0 \\
Identity & 5.5 $\pm$ 0.81 \\
Identity & 5.4 $\pm$ 1.02 \\
Identity & 5.3 $\pm$ 1.68 \\
Inversion with Replacement & 4.9 $\pm$ 0.94\\
Identity & 4.6 $\pm$ 1.2 \\
Inversion & 4.1 $\pm$ 0.83 \\
Inversion with Replacement & 3.6 $\pm$ 0.66 \\
\bottomrule
\end{tabular}
\end{sc}
\end{small}
\end{center}
\vskip -0.1in
\end{table}

\begin{figure}
    \centering
    \scalebox{0.5}{
    \input{avg_reward_plot.pgf}
    }
    \caption{Average reward over 10 training run (with different seeds) of the 10 best hyperparameter/agent indication combinations}
    \label{fig:avg_reward}
\end{figure}